\documentclass[runningheads]{llncs}
\usepackage[T1]{fontenc}

\usepackage{graphicx}
\usepackage{amsmath}
\usepackage{amssymb}
\usepackage{booktabs}
\usepackage{colortbl}
\usepackage{float}
\usepackage{multirow}
\usepackage{array}
\usepackage{placeins}
\usepackage{xcolor}

\usepackage{hyperref}

\usepackage[style=numeric]{biblatex}
\begin{document}
\title{SPARS: Self-Play Adversarial Reinforcement Learning for Segmentation of Liver Tumours}
\titlerunning{Self-play for liver tumour segmentation}

\author{Catalina Tan \inst{1, 2} \and Yipeng Hu \inst{1, 2} \and Shaheer U. Saeed * \inst{1, 2}}
\authorrunning{Tan et al.}
\institute{Department of Medical Physics and Biomedical Engineering, University College London, London, UK\and
 UCL Hawkes Institute, University College London, London, UK\\
 * Email: \email{shaheer.saeed.17@ucl.ac.uk}}
%
% \author{Anonymous authors}
% \institute{Anonymous institute}

\maketitle
\begin{abstract}
Accurate tumour segmentation is vital for various targeted diagnostic and therapeutic procedures for cancer, e.g., planning biopsies or tumour ablations. Manual delineation is extremely labour-intensive, requiring substantial expert time. Fully-supervised machine learning models aim to automate such localisation tasks, but require a large number of costly and often subjective 3D voxel-level labels for training. The high-variance and subjectivity in such labels impacts model generalisability, even when large datasets are available. Histopathology labels may offer more objective labels but the infeasibility of acquiring pixel-level annotations to develop tumour localisation methods based on histology remains challenging in-vivo. In this work, we propose a novel weakly-supervised semantic segmentation framework called SPARS (Self-Play Adversarial Reinforcement Learning for Segmentation), which utilises an object presence classifier, trained on a small number of image-level binary cancer presence labels, to localise cancerous regions on CT scans. Such binary labels of patient-level cancer presence can be sourced more feasibly from biopsies and histopathology reports, enabling a more objective cancer localisation on medical images. Evaluating with real patient data, we observed that SPARS yielded a mean dice score of $77.3 \pm 9.4$, which outperformed other weakly-supervised methods by large margins. This performance was comparable with recent fully-supervised methods that require voxel-level annotations. Our results demonstrate the potential of using SPARS to reduce the need for extensive human-annotated labels to detect cancer in real-world healthcare settings. \\
Code: \url{github.com/catalinatan/SPARS}
\keywords{Cancer \and Reinforcement Learning \and Weak Supervision.}
\end{abstract}

\section{Introduction}

Tumour segmentation is crucial for diagnosing and treating liver cancer, particularly in its early stages \cite{2009, LiTS, 5, 6, 7}. For instance, identifying cancerous lesions from medical scans can enable clinicians to estimate the tumour diameters and volumes required for targeted radiation delivery in radiotherapies, where radiation dose varies with tumour size \cite{5}. Currently, radiologists manually delineate tumour boundaries on each slice of 3D computed tomography (CT) and/ or magnetic resonance imaging (MRI) scans \cite{2009}. However, this manual delineation may be subjective, time-consuming and poorly reproducible \cite{9, 10, 11}. The subjectivity leads to high inter- and intra-clinician variability in such tasks, which may be a result of the morphological diversity of tumour appearances on these scans as well as varying institute-specific training and expertise \cite{9,10}. 
% These problems underscore the need for viable alternatives like automated segmentation to optimise tumour localisation for cancer interventions \cite{14, 15}.
These problems underscore the need for more reproducible and objective tumour localisation for cancer interventions \cite{14, 15}.

Automated segmentation methods aim to address the reproducibility of tumour localisation using fully-supervised learning, where a large number of labels that may be curated through consensus of multiple experts are used to train an automated model. However, this requires a large number of voxel-level expert-annotated labels that are very costly to obtain, especially if curated with a consensus from multiple experts \cite{SIPE, 2009, ACR, saeed2024active}. 

To mitigate the challenges that plague fully-supervised learning, weakly-supervised learning has been explored for a variety of tasks. Weak supervision allows the use of weak labels (during training or model development) to perform complex tasks at inference. Examples include training neural networks to perform pixel-level segmentation using only bounding boxes \cite{bounding1, bounding2}, scribbles \cite{scribble1,scribble2} or image-level annotations \cite{image-level1, image-level2, ShaheerFSP, 26, yi2023boundary} during training. In particular, image-level classification labels present great promise for weakly-supervised semantic segmentation (WSSS) due to the comparatively lower cost of their acquisition \cite{ACR}. However, achieving comparable performance to fully-supervised learning, especially in complex clinical tasks such as tumour localisation, remains an open challenge. 
% likely due to a lack of tunable application-specific hyper-parameters, which can have large impact on performance, especially in complex medical imaging tasks \cite{nn-UNet}.

In this work we propose a novel framework for WSSS, where we use a minimal number of image-level labels of cancer presence during model development to allow a voxel-level tumour segmentation at inference. The image-level cancer presence labels are objective histopathology labels which indicate whether each patient has clinically significant cancer or not. 
% We use self-play adversarial reinforcement learning (RL) where an agent competes against a historical version of itself to localise regions-of-interest (ROI) by moving a window of a pre-defined size across an image. A pre-trained object presence classifier provides a score to indicate the likelihood of ROI-presence within the window, where the classifier is itself trained using only image-level binary object presence labels. If the agent receives a higher score than its opponent, it wins the turn and receives a positive reward signal. By favouring movements that lead to positive rewards, the agent learns to move its observation window towards potential ROIs. Both agents continue to compete against each other until one reaches a termination signal which indicates that a sufficient localisation accuracy has been reached. This is one of the thresholds that can be tuned to achieve optimal segmentation performance for the task-of-interest. 
These image-level labels are used to train an object presence classifier which can classify ROI (region-of-interest) presence within an image. At inference, we use this classifier to generate logits (classification probability) for a section or window of the image, which serves as a likelihood of object presence within the window. For localisation of ROIs, we use self-play adversarial reinforcement learning (RL) where two agents compete to localise ROIs. Each agent moves a window across an image and is rewarded based on the classifier output for the window indicating object presence likelihood. Training to maximise the reward allows each agent to improve localisation towards areas where the likelihood of object presence is maximised. 
Tracking the classifier outputs for each window, we can generate a voxel-level probability map (described in Sec. \ref{sec:methods_seg}), where a threshold can control which voxels are to be included as positive or negative for the final segmentation. This allows greater application-specific flexibility compared to other weakly-supervised methods, with fixed localisation termination conditions \cite{ShaheerFSP, yi2023boundary, 26}. 

The contributions of our work are summarised: \\
\begin{enumerate}
    \item we propose a self-play adversarial RL framework for WSSS to minimise manual segmentation costs;
    \item we propose to supervise the adversarial RL framework using a classifier trained on image-level labels of object presence to quantify ROI presence likelihood which forms the rewards for RL-based WSSS;
    \item our method allows greater application-specific flexibility compared to other WSSS methods by generating a pixel-level classification map (as opposed to patch-level \cite{ShaheerFSP, 26}) which can be thresholded to adjust which pixels to include as positive or negative 
    \item we evaluate our method using data from real liver cancer patients, to localise tumours on CT scans, and compare with recent state-of-the-art weakly- and fully-supervised algorithms demonstrating superior performance to weakly-supervised, and comparable performance to fully-supervised, approaches;
    \item we make our algorithm openly available: \url{github.com/catalinatan/SPARS}
\end{enumerate}

\section{Methods}

\begin{figure}
    \centering
    \includegraphics[width=0.98\textwidth]{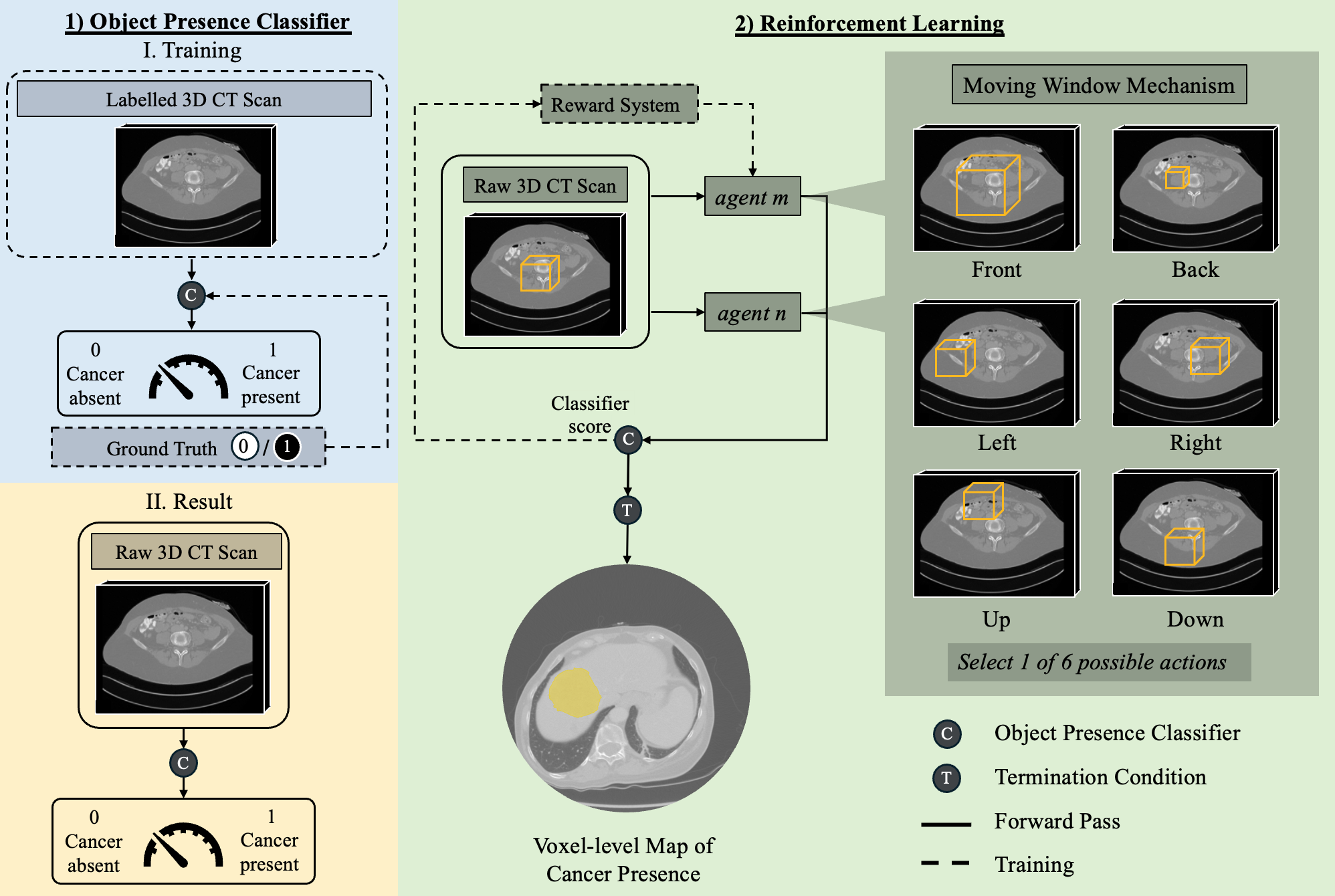}
    \caption{An overview of the proposed method, where two agents compete to localise ROIs guided by a classifier trained only using weak classification labels.}
    \label{fig:enter-label}
\end{figure}

\subsection{Object Presence Classifier}\label{sec:classifier}

The object presence classifier $f(\cdot; w): \mathcal{X} \rightarrow[0,1]$, generates a score for a given image sample $x \in \mathcal{X}$ where $\mathcal{X}$ is the image domain and $w$ represents the neural network parameters. This is modelled as a mechanism to provide a linkage function between the two differing weak and full objectives, here, image-level and pixel-level classifications \cite{26, ShaheerFSP, CAM1, CAM2, saeed2025guided}. These network parameters are optimised using image-label pairs which indicate object presence in the sample, denoted as \(\left\{x_{i}, y_{i}\right\}_{i=1}^{N}\) where $y_i \in \mathcal{Y}$ with $\mathcal{Y}$ being the label domain $\{0,1\}$ and $N$ is the number of samples in the set. If the region of interest (ROI) is contained in $x_i$, its corresponding binary label is given by \(y_{i}=1\). Conversely, \(y_{i}=0\) if the ROI is not present. 

The classifier is trained using the binary cross-entropy loss function: 

\[l(y_i,f(x_i;w))= - \frac{1}{N} \sum^{N}_{i=1}(y_i \space \log(f(x_i;w))+(1-y_i) \space \log(1-f(x_i;w))\]

where $f(x_i;w)$ is the predicted label by the classifier and $y_i$ is the ground truth label for an image $x_i$. 

The network parameters $w$ are optimised by minimising the expected loss function: 
\[w^\ast = \arg\min_w~ \mathbb{E}_{x_i\in\mathcal{X}, y_i\in\mathcal{Y}} [l(y_i,f(x_i;w)]\]
where \(w^\ast\) represents the optimal parameters.

\subsection{Markov decision process environment}

In RL, an agent (neural network) interacts with an environment by producing actions. In response to agent actions, the environment generates new states that are observed by the agent, and rewards that inform the agent of the impact of its actions. The rewards are used to train the agent to predict optimal actions given states. The agent-environment interactions are modelled as a Markov decision process that iterates sequentially, and is represented by a tuple $(\mathcal{S},\mathcal{A},p,r,\pi, \gamma)$, where $\mathcal{S}$ denotes the state space,  $\mathcal{A}$ represents the action space, $p$ defines the state transition probability, $r$ is the reward, $\pi$ is the agent or policy, and $\gamma$ represents the discount factor for future rewards.

\paragraph{}\textbf{States:} The observed states are cropped windows $x_i^{(a, b, c)}$, where $x_i$ is the entire image and $(a, b, c)$ denote the locations of a fixed-size crop in the height, width and depth dimensions respectively (a hyper-parameter investigated in Sec. \ref{sec:res_train_size}). The state $s_t\in\mathcal{S}$ at time-step $t$ is therefore given by $s_t = \{ x_i^{(a, b, c)}\}$, where $\mathcal{S}$ denotes the state space. 

\paragraph{}\textbf{Actions:} The actions can move the window $x_i^{(a, b, c)}$ in any direction. The action $a_t\in\mathcal{A}$ is given by $a_t = (\delta a, \delta b, \delta c)$. The actions are modelled discretely, only allowing a movement by a fixed distance in one of 6 directions, which means that 
$(\delta a, \delta b, \delta c) \in \{(\pm d, 0, 0), (0, \pm d, 0), (0, 0, \pm d)\}$, where only one of \(\delta a\), \(\delta b\), or \(\delta c\) can be non-zero (i.e., \(\pm d\)), while the others must be zero. 

\paragraph{}\textbf{State transitions:} Given the current state $s_t = \{ x_i^{(a, b, c)}\}$ and action $a_t = (\delta a, \delta b, \delta c)$, the next state $s_{t+1}= \{ x_i^{(a+\delta a,~ b+\delta b,~ c+\delta c)}\}$. So the action moves the window to a new location. The probability of transitioning to the next state $s_{t+1}$, given current state $s_t$ and action $a_t$ is given by $p(\cdot): \mathcal{S} \times \mathcal{S}\times \mathcal{A} \rightarrow [0,1]$, which can be expressed as $p(s_{t+1}|s_t,a_t)$ for a particular time-step $t$.

\paragraph{}\textbf{Policy:} The policy or agent $\pi(\cdot; \theta): \mathcal{S} \rightarrow \mathcal{A}$ predicts the action $a_t$, given the state $s_t$, where $\theta$ represents the parameters of the policy and $a_t = \pi(s_t; \theta)$. Note that in this work we model the policy stochastically where $\pi(\cdot;\theta): \mathcal{S} \times \mathcal{A} \rightarrow [0,1]$, which gives the probability whereby $a_t \sim \pi(\cdot | s_t)$, however, the notation of $a_t = \pi(s_t; \theta)$ is adopted for simplicity.

\subsection{Policy optimisation using self-play adversarial RL}

The use of two competing agents to localise ROIs ensures that the impact of any errors in the object presence classifier are minimised since rewards for training are based on a comparison rather than exact object presence classifier outputs. Furthermore, using experience from two agents allows convergence with fewer time-steps compared to using only one set of experiences for training. Previous works have also shown the efficacy of using self-play in object localisation \cite{ShaheerFSP}.

In our work we use self-play adversarial RL, where there are two versions of the policy, denoted as $\pi_m(\cdot; \theta_m)$ and $\pi_n(\cdot; \theta_n)$, where the only difference is the parameters. In further analyses subscripts $m$ and $n$ will denote outputs from polices $m$ and $n$ respectively. In this work, $\theta_m = \theta_n$. Since the policy is modelled stochastically, the actions selected by each may differ at any single time-step.

\paragraph{}\textbf{Rewards:} A reward function $r(\cdot): \mathcal{S} \times \mathcal{A} \rightarrow \mathbb{R}$ predicts a reward $R_t \in \mathbb{R}$, given the state $s_t$  and action $a_t$. For policy $\pi_m$, the state, action, reward triplet is given by $(s_{m, t}, a_{m, t}, R_{m, t})$ and for policy $\pi_n$ by  $(s_{n, t}, a_{n, t}, R_{n, t})$. The reward for $\pi_m$ is given by:

$$
R_{m, t} =
\begin{cases}
+1 & \text{if } f(s_{m, t}; w^*) \geq f(s_{n, t}; w^*) \\
-1 & \text{otherwise}
\end{cases}
$$

This means that a positive reward is given for policy $\pi_m$ when the likelihood of object presence as measured by the trained object presence classifier for the window $s_{m, t}$, given by $f(s_{m, t}; w^*)$, is greater than the likelihood of object presence for the window $s_{n, t}$, given by $f(s_{n, t}; w^*)$. 

The reward for $\pi_n$ is just the opposite of reward for $\pi_m$ and is given by $R_{n, t} = - R_{m, t}$.

Despite training of the object presence classifier using only image-level labels (as outlined in Sec. \ref{sec:classifier}), it can be used for inference on windows, where the windows can be resized to the original image size for classifier inference, similar to previous work \cite{ShaheerFSP, 26, yi2023boundary}. This serves as a linkage function which links image-level labels to windows, despite training only with image-level labels. The difference in appearances between images and windows may impact classifier likelihood predictions. However, the competing mechanism described above ensures that such impact is minimised as rewards for training are only based on a comparison of window-level likelihoods rather than exact values.

\paragraph{}\textbf{Policy optimisation:} The state, action, reward triplets are collected for a total of $T$ time-steps for both policies, giving $\tau_m = (s_{m,0},a_{m,0},R_{m,0}, \dots, s_{m,T},a_{m,T},R_{m,T})$ and $\tau_n = (s_{n,0},a_{n,0},R_{n,0}, \dots, s_{n,T},a_{n,T},R_{n,T})$. The return over these time-steps is given by $R(\tau) = \sum_{k=0}^T \gamma^k R_{t+k}$. Where $R(\tau_m)$ and $R(\tau_n)$ denote the returns for policy $\pi_m$ and $\pi_n$ respectively. The optimisation problem is formulated as:

$$
\theta^* = \arg\max_\theta \{ \mathbb{E}_{\tau_m} [R(\tau_m)] +  \mathbb{E}_{\tau_n} [R(\tau_n)]\}
$$

\subsection{Segmentation of new samples}\label{sec:methods_seg}

The optimised policy $\pi(\cdot; \theta^*)$ can then be used to conduct the segmentation for a new sample $x_n$. The state starts out as a window in the centre of the image $s_0 = x_n^{(a_0,b_0,c_0)}$ and following the optimised policy $\pi(\cdot; \theta^*)$ produces a trajectory of states and actions $(s_0, a_0, ..., s_E,a_E)$, where $E$ denotes the iteration at which the object presence classifier reaches a threshold $f(x_n^{a_E, b_E, c_E}; w^*)>\rho$, where $\rho$ is the threshold that ensures localisation has reached a sufficient accuracy (configured through a grid search as outlined in experiments). The segmentation map is assembled using object presence classifier outputs for each window $(f(x_n^{a_0, b_0, c_0}; w^*), .., f(x_n^{a_E, b_E, c_E}; w^*))$ where the voxel-level probabilities for each window are denoted by $f(x_n^{a_t, b_t, c_t}; w^*)$. A full segmentation map $z_n$ starts out with all voxel-level probabilities being 0, the probability values for a window in the segmentation map $z_n^{a_t, b_t, c_t}$ is updated as $z_n^{a_t, b_t, c_t} \leftarrow z_n^{a_t, b_t, c_t} + f(x_n^{a_t, b_t, c_t}; w^*)$. Starting with voxel-level probabilities being 0 allows the agents to minimise placing windows in locations where the ROI is extremely unlikely to exist as these voxel-level probabilities can be left unchanged throughout the localisation process. Accumulating the voxel-level probabilities for all time-steps until $E$, we get the final voxel-level probability map, which can be thresholded to obtain the final segmentation. The threshold for obtaining the final segmentation can be optimised per-application, which allows additional flexibility.

\section{Experiments}

\subsection{Dataset}

In our experiments we use 131 portal venous phase 3D computed tomography (CT) scans from diagnosed patients in the Liver Tumour Segmentation Benchmark (LiTS) dataset, each depicting zero to twelve clinically significant primary and secondary liver tumours \cite{LiTS}. Each scan is accompanied by voxel-level annotations produced by experienced radiologists with values 0 (non-hepatic tissue), 1 (non-cancerous hepatic tissue) and 2 (cancerous hepatic tissue) as ground truth labels. Binary image-level labels (0 for non-cancerous tissue and 1 for cancerous tissue) were derived from these voxel-level labels during training and testing. The development-to-testing ratio for this dataset was 3:2.

\subsection{Model architectures}

The object presence classifier $f$ consists of 4 convolutional blocks (3D convolutional layer with a $(3 \times 3 \times 3)$ kernel, 3D batch normalisation and rectified linear unit (ReLu) followed by 3D max pooling with a $(2\times 2\times 2)$ kernel) and 5 fully connected layers (with ReLU activations).

The RL policy follows the same architecture, with the only change being in the final output layer. The policy optimisation \cite{PPO} uses only the experience from $\pi_m$ for training in our experiments.

For other hyper-parameter settings please refer to experiments (specified as appropriate) or code available in open-source repository.

The classifier took approximately 12 hours to train, and RL took approximately 96 hours, on a single Nvidia Tesla V100 GPU. The classifier inference time on the same hardware was 126ms and the RL inference time to obtain the final segmentation was 1.8s on average.

\subsection{Experimental protocol and comparisons}

\subsubsection{Ablations:}
In our framework, we evaluate the impact of the training dataset size for the object presence classifier as well as the impact of hyper-parameters including the window size and threshold $\rho$ which controls the termination of the final segmentation when a sufficient accuracy has been reached.

\subsubsection{Comparisons:}
We compare our method with recent state-of-the-art weakly-supervised methods based on multi-instance learning \cite{SA-MIL}, class-activation maps \cite{ACR}, RL \cite{ShaheerFSP} and region-classification \cite{MARS,MCT, SIPE} for the same task on the same dataset and also with recent state-of-the-art fully-supervised methods including recent state-of-the-art for the task \cite{CLIP} and other common baselines \cite{nn-UNet, 21, 22}.

\section{Results}

\subsection{Ablations}

\subsubsection{Training set size for object presence classifier:}\label{sec:res_train_size}
\begin{figure}[H]
    \centering
    \includegraphics[width=0.75\linewidth]{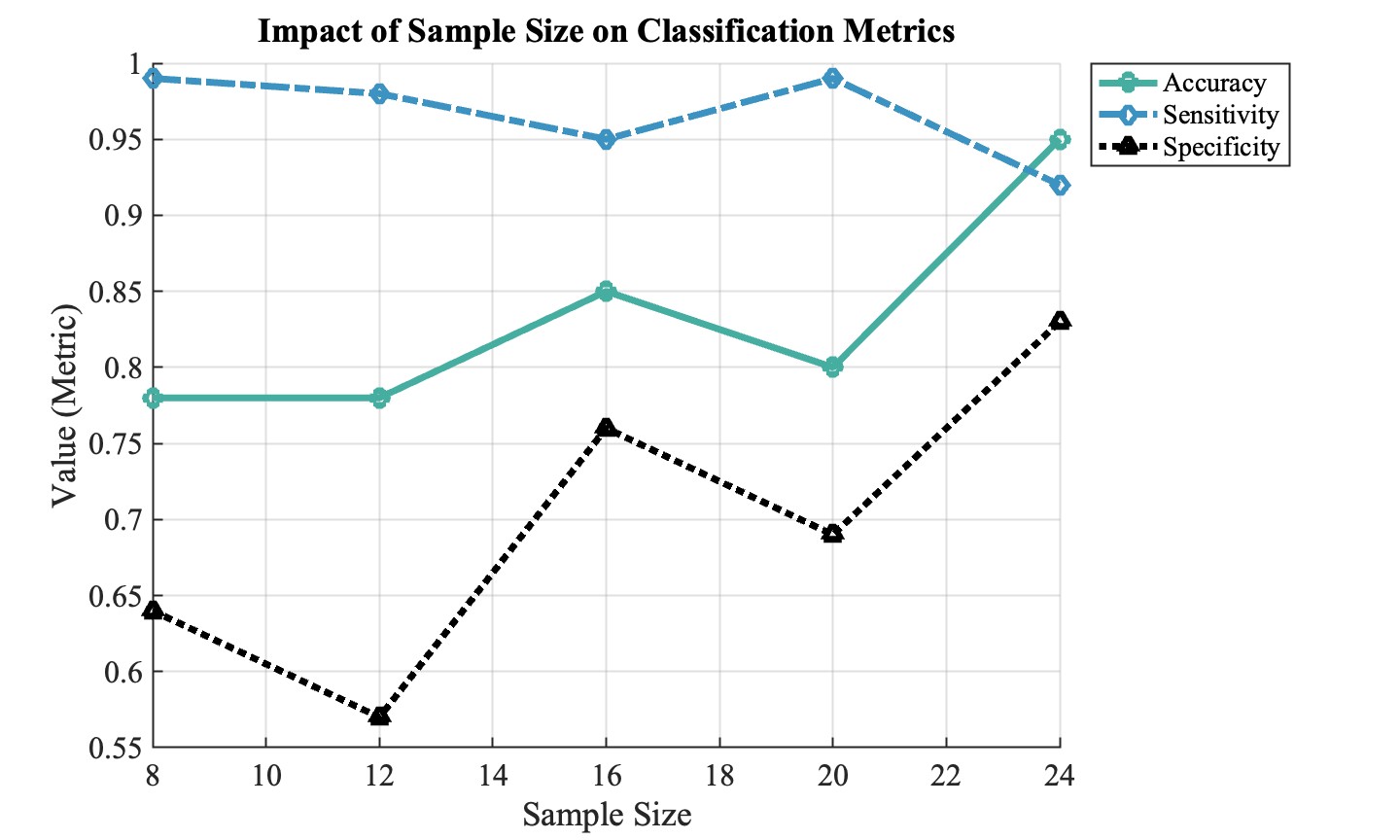}
        \caption{Training set size against segmentation performance.}
    \label{fig:CNN-classifier}
\end{figure}

Fig. \ref{fig:CNN-classifier}, increasing sample size leads to a positive trend in the four performance metrics overall. This result aligns with those of Althian et al. \cite{dataset-size}, which found that their neural network performance metrics decrease with smaller training datasets. In terms of the network's true positive and negative rates, the algorithm's specificity is consistently lower than its sensitivity regardless of sample size, which suggests that it may over-predict positive cancer classes. 

\subsubsection{Impact of window size:} 
\begin{figure}[H]
    \centering
    \includegraphics[width=0.75\linewidth]{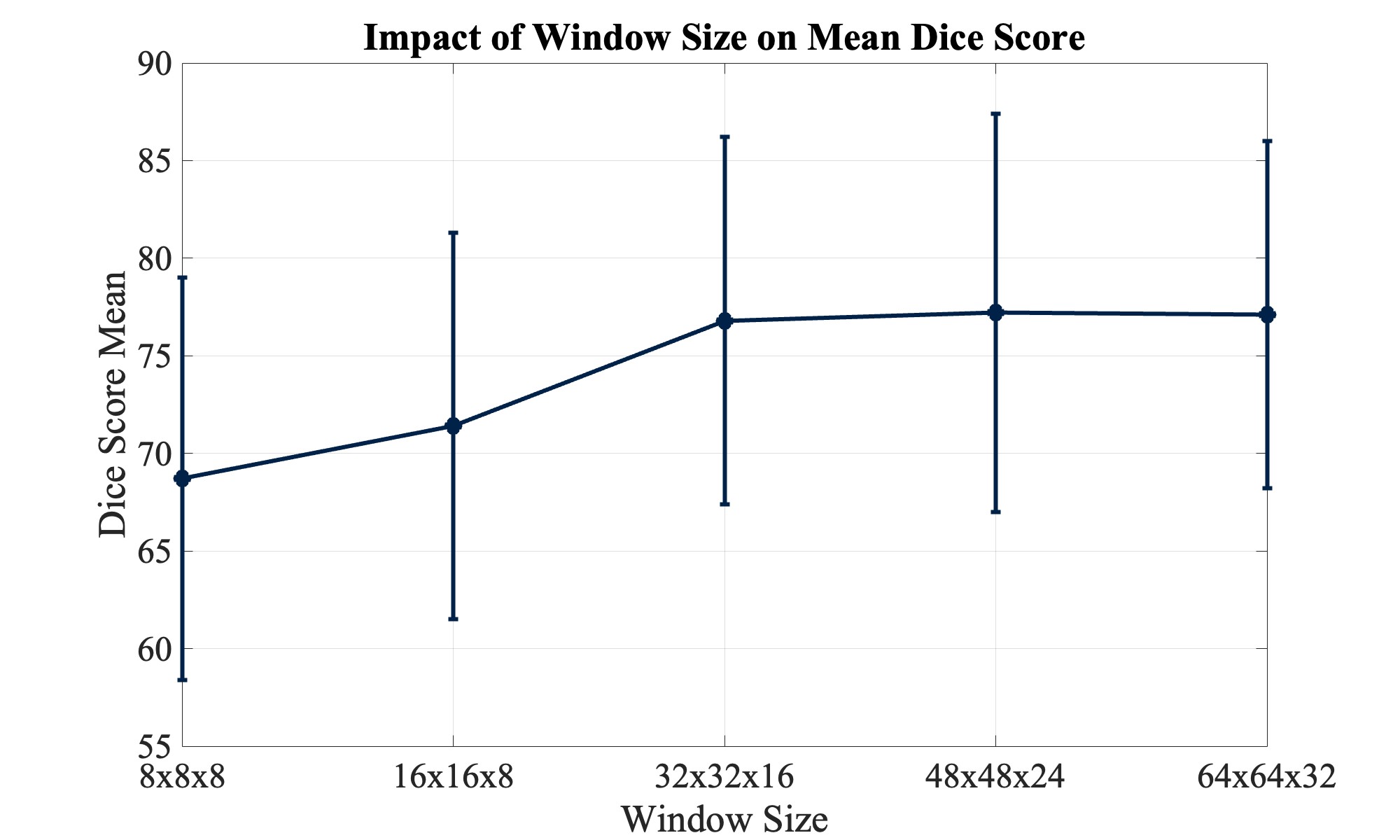}
    \caption{Window size against segmentation performance.}
    \label{fig:RL-window}
\end{figure}

Fig. \ref{fig:RL-window} shows that increasing the size of the window (state observed by an agent), tends to result in an increase in the mean dice score for the final segmentation, until it reaches a plateau after a window size of $32 \times 32 \times 16$.  After this point, the mean dice score tends to stabilise. 

\subsubsection{Impact of termination threshold:} 
\begin{figure}[H]
    \centering
    \includegraphics[width=0.75\linewidth]{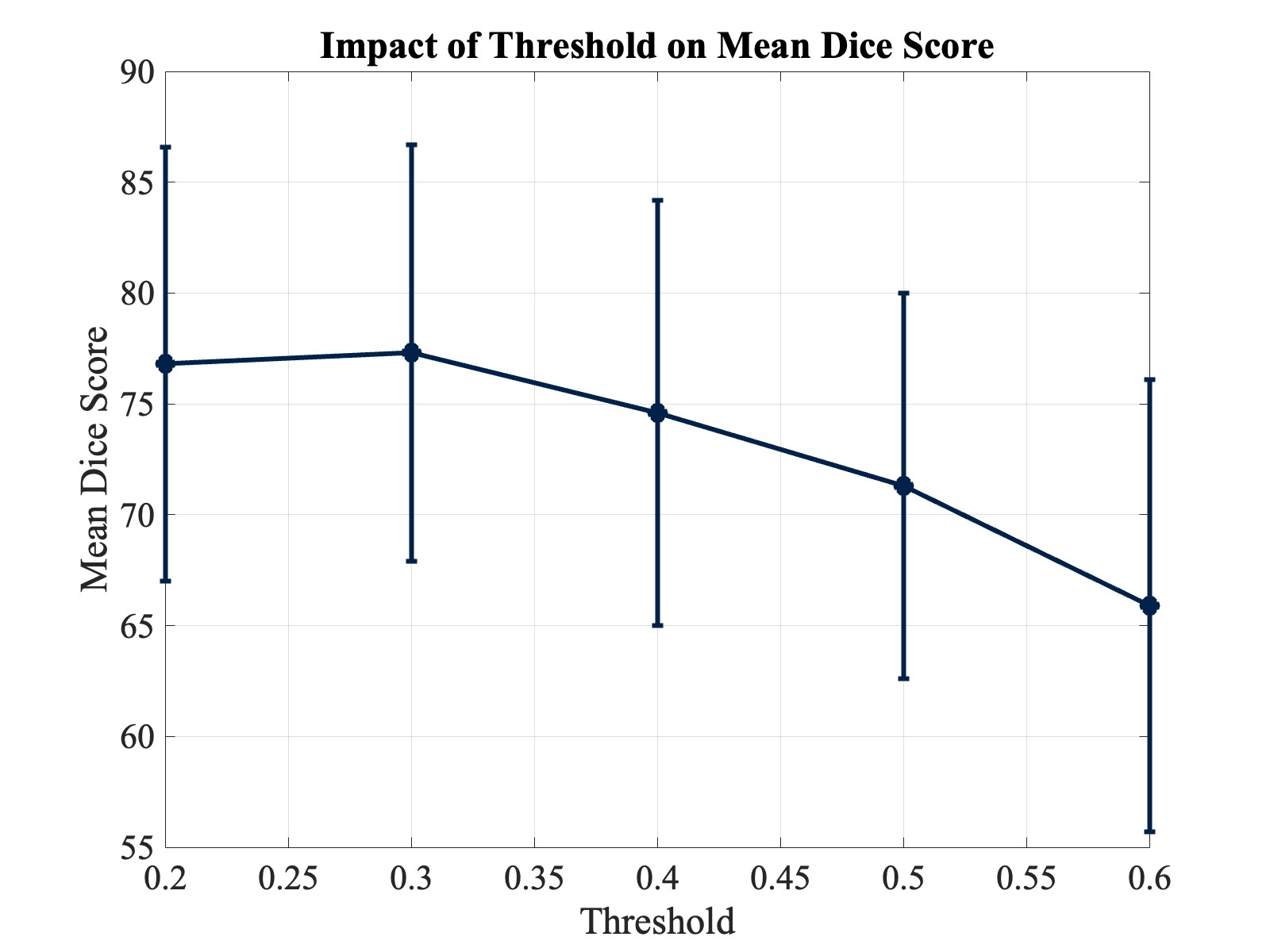}
        \caption{Threshold against segmentation performance.}
    \label{fig:RL-threshold}
\end{figure}

Fig. \ref{fig:RL-threshold} shows that increasing the threshold that controls termination leads to an increase in mean dice score for the final segmentation, until 0.3, where the mean dice score begins to decline. This suggests that the output from the object presence classifier at a single time-step may be relatively low $f(s_{t_a};w^*), f(s_{t_b};w^*) \leq 0.3$. As a result, final predicted classifications primarily rely on accumulated predictions over a trajectory instead of over a single time-step. 

\subsection{Comparisons}

\subsubsection{Comparisons with weakly-supervised methods:}

\begin{figure}[H]
    \centering
    \includegraphics[width=0.75\linewidth]{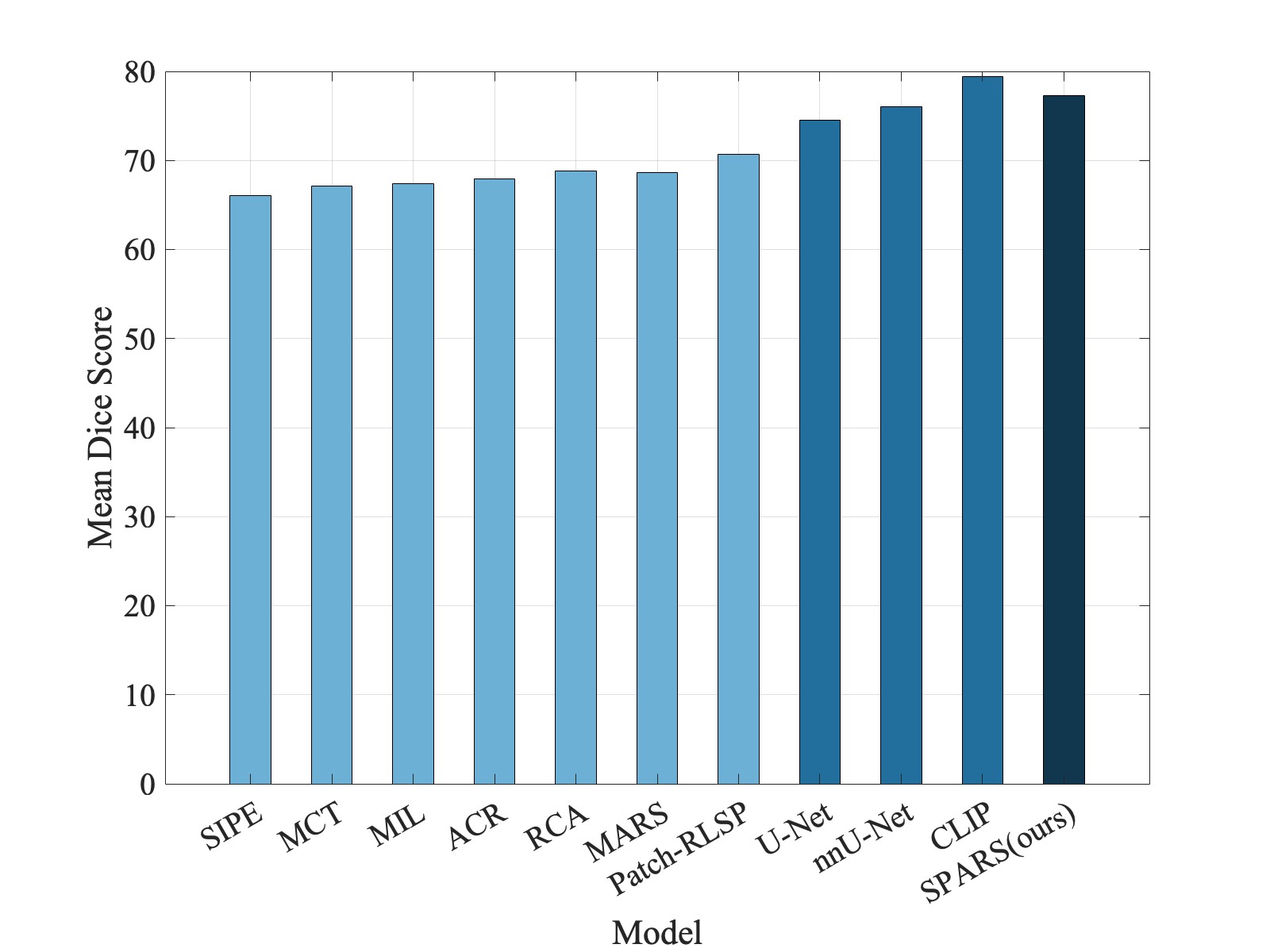}
    \caption{Performance compared to other models for liver tumour segmentation (U-Net, nnU-Net and CLIP are fully-supervised and rest are all weakly-supervised).}
    \label{fig:method-compare}
\end{figure}

Tab. \ref{tab:method-dice} and Fig. \ref{fig:method-compare} show that our approach yielded in a mean dice score (Dice) 6.6 percentage points higher than the previous state-of-the-art method \cite{ShaheerFSP}. The Dice of 77.3 achieved by our method not only outperforms the tested WSSS algorithms but aligns with recent fully-supervised methods. Statistical tests were not conducted due to the absence of standard deviations on reported dice scores for other methods. 
\begin{table}[H]
    \centering
    \caption{Performance compared to other models for liver tumour segmentation. (Adapted from \cite{ShaheerFSP})}
    \begin{tabular}{cccc}
        \toprule
        \textbf{Method} & \textbf{Supervision} & \textbf{Dice} & mIoU\\
        \midrule
        SIPE \cite{SIPE} & Weak & 66.1  & 49.3\\
        MCT \cite{MCT} & Weak & 67.1    & 50.5\\
        MIL \cite{SA-MIL}& Weak & 67.4  & 50.8\\
        ACR \cite{ACR} & Weak & 67.9  & 51.4\\
        RCA \cite{RCA} & Weak & 68.8    & 52.4\\
        MARS \cite{MARS} & Weak & 68.6  & 52.2\\
        Patch-RLSP \cite{ShaheerFSP}& Weak & 70.7 & 54.7\\
        \hline
        U-Net \cite{21} & Full & 74.5       & 59.3\\
        nnU-Net \cite{nn-UNet}& Full & 76.0 & 61.3\\
        CLIP \cite{CLIP} & Full & 79.4      & 65.8\\
        \hline
        SPARS(ours)& Weak & 77.3            &62.9\\
        \bottomrule
    \end{tabular}

    \vspace{0.1cm}

    \centering
    \label{tab:method-dice}
\end{table}

\subsubsection{Comparisons with fully-supervised methods:}
Tab. \ref{tab:method-dice} and Fig. \ref{fig:method-compare} show that our approach outperformed several existing fully-supervised models including U-Net ($\text{Dice}=74.5$) and performed comparably to CLIP which is the current state-of-the-art for the tested task ($\text{Dice}=79.4$) \cite{21}, where our method only had 2.5\% lower Dice ($\text{Dice}=77.3$). However, the methods compared against, used the full segmentation labels for training which are much more costly to obtain compared to weak labels that are used in our method. Additionally, we only used 24 weak labels for training compared to over 64 full segmentation labels used by other methods. Statistical tests were not conducted due to the absence of standard deviations on reported dice scores for other methods. 

\subsection{Qualitative results}
\begin{figure}
    \centering
    \includegraphics[width=1\linewidth]{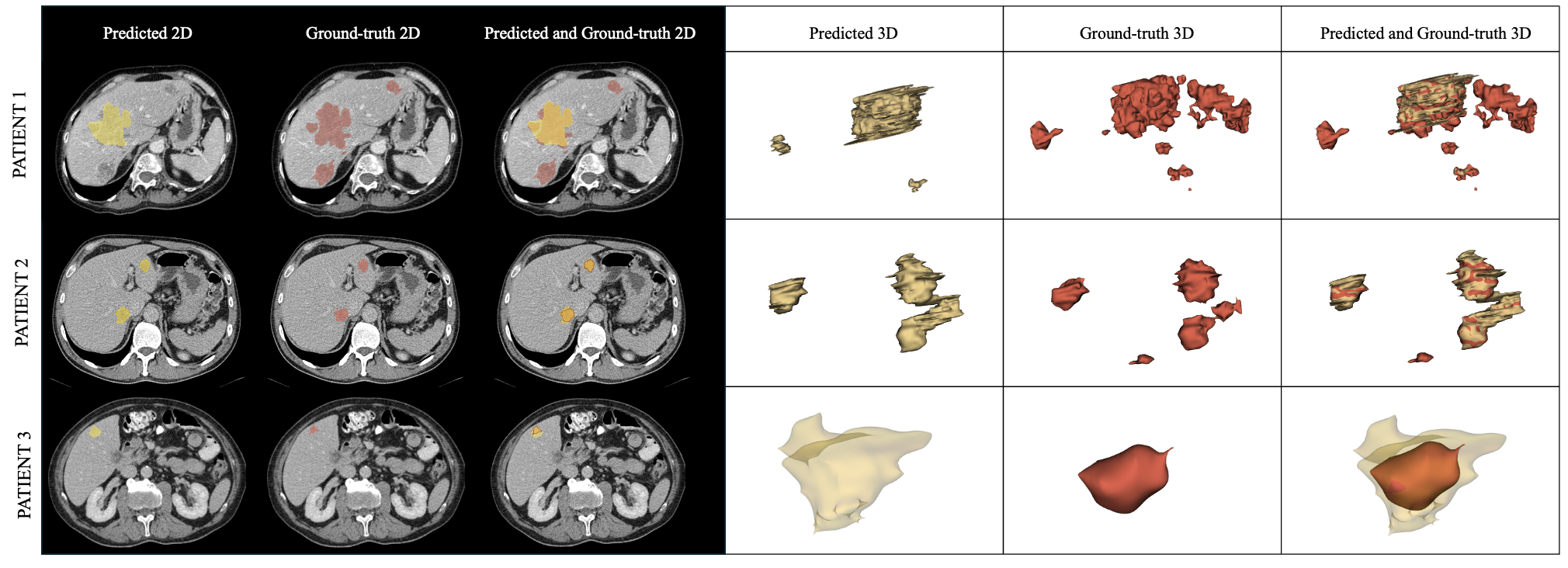}
    \caption{Qualitative examples for a comparison between our predicted segmentations and ground-truth from radiologist annotations.}
    \label{fig:qualitative_results}
\end{figure}

Fig. \ref{fig:qualitative_results} shows qualitative examples of predicted and ground truth liver tumour segmentations on patient data. We observed that larger tumours (volume $>10^3 ~\text{mm}^3$) were under-segmented (Patient 1) compared to smaller tumours (Patients 2 and 3). We also observed a 24.5\% lower average Dice for small tumours compared to their larger counterparts. Luan et al \cite{Luan2021} reported similar trends for their liver tumour segmentation framework where smaller tumours (< 0.2 cm) achieved a 60\% lower mean dice score ($\text{Dice}=0.32$) than larger tumours.

\section{Discussion}

We observed a large performance improvement for our method compared to the previous best-performing weakly-supervised method \cite{ShaheerFSP} for the same application. Despite both using self-play RL for WSSS, our method enables voxel level probability maps by accumulating the classifier predictions using a moving window (summarised in Sec. \ref{sec:methods_seg}). This enables multiple passes over the same areas to increase segmentation confidence unlike previous patch-based approaches that only allow binary selection or rejection of patches or pixels \cite{ShaheerFSP}. This, along with the novel termination condition based on a thresholded classifier score, mean that our framework allows more flexibility in tuning thresholds and parameters for different applications. This application-specific tuning in our flexible framework may be the reason that we observed superior performance compared to recent state-of-the-art methods.

The ablation studies reveal the impact of key hyper-parameters on performance. The performance plateaus after a large enough window size. The performance improvement observed with increasing window size may be due to the fact that as windows become larger their appearance starts to match the appearance of full images. The performance peaks at termination threshold of 0.3, indicating that low thresholds may not be suitable as they may terminate segmentation too early (under-prediction) whereas high thresholds may terminate segmentation too late (over-prediction). The skew of the optimal threshold towards under-prediction may be because of an over representation of cancer-positive classes in the dataset.

Our method performed comparably with fully-supervised approaches despite only using weak labels during training, which are much more cost-effective to obtain compared to voxel-level segmentation labels. This highlights that cost-effective training mechanisms like WSSS may be viable alternatives to fully-supervised learning when data is constrained or of poor quality.

Future investigation could explore alternative reward or termination conditions that allow greater flexibility for application-specific tuning, which may allow even further performance improvements.

\section{Conclusion}

In this work, we proposed self-play adversarial learning for semantic segmentation (SPARS) where agents compete to localise ROIs. Only weakly supervised labels were utilised to train an object presence classifier, which guides and scores the competition, allowing agents to move windows closer to ROIs. This approach successfully outperformed existing state-of-the-art WSSS models and performed comparably to fully-supervised method which used costly voxel-level segmentations for training. This highlights the cost-effective nature of weakly-supervised training with minimal compromise on performance compared to costly fully-supervised training.

\section*{Acknowledgements}

This work is supported by the International Alliance for Cancer Early Detection, an alliance between Cancer Research UK [EDDAPA-2024/100014] \& [C73666/A31378], Canary Center at Stanford University, the University of Cambridge, OHSU Knight Cancer Institute, University College London and the University of Manchester; and National Institute for Health Research University College London Hospitals Biomedical Research Centre.

\printbibliography

\end{document}